\pdfoutput=1

\documentclass[11pt]{article}

\PassOptionsToPackage{sort}{natbib}

\usepackage{ACL2023}
\usepackage{booktabs}
\usepackage[inline]{enumitem}
\usepackage[skip=2pt]{caption}
\usepackage{acronym}
\usepackage{multirow}
\usepackage{makecell}
\usepackage{graphicx}
\usepackage{times}
\usepackage{xcolor}
\usepackage{subcaption}
\usepackage{wrapfig}
\usepackage{amsmath}
\usepackage{xspace}

\usepackage{times}
\usepackage{latexsym}

\usepackage[T1]{fontenc}

\usepackage[utf8]{inputenc}

\usepackage{microtype}

\usepackage{inconsolata}

\usepackage{amssymb}
\usepackage{algorithm}
\usepackage{algpseudocode}

\usepackage{tikz}
\usetikzlibrary{shapes.geometric, arrows, positioning, fit, calc, patterns}
\usepackage{graphicx}

\usepackage{newfloat}
\DeclareFloatingEnvironment[fileext=lop, listname={List of prompts}, 
                            name=Prompt, placement=h]{prompt}

\usepackage{listings}
\usepackage{xcolor}
\AtBeginDocument{%
  \providecommand\BibTeX{{%
    \normalfont B\kern-0.5em{\scshape i\kern-0.25em b}\kern-0.8em\TeX}}}

\acrodef{TDS}{task oriented dialogue system}
\acrodef{AMT}{amazon mechanical turk}
\acrodef{ReDial}{recommendation dialogues}
\acrodef{LLM}{large language model}
\acrodef{IR}{information retrieval}
\acrodef{CRS}{conversational recommender system}
\acrodef{HIT}{human intelligence task}
\acrodef{ICC}{intraclass correlation coefficient}
\acrodef{KDE}{kernel density estimation}
\acrodef{CIS}{conversational information-seeking}
\acrodef{CS}{conversational search}

\newcommand{\header}[1]{\vspace{1mm}\noindent\textbf{#1.}}

\allowdisplaybreaks

\author{
    Clemencia Siro‡ Yifei Yuan§ Mohammad Aliannejadi‡ Maarten de Rijke‡ \\
    ‡University of Amsterdam \\
    §University of Copenhagen \\
    ‡\texttt{\{c.n.siro, m.aliannejadi, m.derijke\}@uva.nl} \\
    §\texttt{yiya@di.ku.dk}
}

\begin{document}

\title{AGENT-CQ: Automatic Generation and Evaluation of Clarifying Questions for Conversational Search with LLMs}

\maketitle

\begin{abstract}
Generating diverse and effective clarifying questions is crucial for improving query understanding and retrieval performance in open-domain \acf{CS} systems. 
We propose \textbf{AGENT-CQ} (\textbf{A}utomatic \textbf{GEN}eration, and evalua\textbf{T}ion of \textbf{C}larifying \textbf{Q}uestions), an end-to-end LLM-based framework addressing the challenges of scalability and adaptability faced by existing methods that rely on manual curation or template-based approaches. AGENT-CQ consists of two stages: \emph{a generation stage}  employing LLM prompting strategies to generate clarifying questions, and \emph{an evaluation stage (CrowdLLM)} that simulates human crowdsourcing judgments using multiple LLM instances to assess generated questions and answers based on comprehensive quality metrics. Extensive experiments on the ClariQ dataset demonstrate CrowdLLM's effectiveness in evaluating question and answer quality. Human evaluation and CrowdLLM show that AGENT-CQ -- generation stage, consistently outperforms baselines in various aspects of question and answer quality. In retrieval-based evaluation, LLM-generated questions significantly enhance retrieval effectiveness for both BM25 and cross-encoder models compared to human-generated questions. We will make publicly available the data, labels and prompts used by AGENT-CQ framework. \looseness=-1

\end{abstract}

\acresetall

\section{Introduction}

\Acf{CS} systems have gained significant attention in recent years, offering users a more natural and interactive way to find information than single-shot search interactions~\citep{DBLP:conf/chiir/RadlinskiC17,AliannejadiZCC19,DBLP:journals/ftir/ZamaniTDR23}. 
To resolve the ambiguity inherent in user queries, these systems may ask users clarifying questions~\cite{DBLP:conf/sigir/ZamaniMCLDBCD20}. 
Generating diverse and effective clarifying questions is crucial for improving query understanding and retrieval performance, which remains a challenge~\cite{DBLP:conf/www/0002SARL24}.

\begin{figure}[!t]
\centering
\resizebox{\linewidth}{!}{%
\begin{tikzpicture}[
scale=0.65,
    transform shape,
    node distance=1.5cm and 0.7cm, %
    realdata/.style={
        cylinder, draw, cylinder uses custom fill,
        shape border rotate=90, aspect=0.25,
        minimum width=2.5cm, minimum height=1cm,
        text width=2.3cm, align=center, font=\small
    },
    generateddata/.style={
        cylinder, draw, shape border rotate=90, aspect=0.25,
        minimum width=2.5cm, minimum height=1cm,
        text width=2.3cm, align=center, font=\small
    },
    process/.style={
        rectangle, draw, fill=white,
        minimum width=2.5cm, minimum height=1cm,
        text centered, font=\small
    },
    highlight/.style={
        rectangle, draw, fill=#1,
        minimum width=2.5cm, minimum height=1cm,
        text centered, font=\small
    },
    phase/.style={
        rectangle, draw=gray, dashed,
        inner sep=0.3cm, rounded corners
    },
    arrowlabel/.style={
        midway, fill=white, font=\small,
        text centered
    },
    verticaltext/.style={
        rotate=90, anchor=west
    }
]

\draw[black, rounded corners=2, thick] (-1.7,-7.4) rectangle (11,2.5) ;

    \node at (9.38,-7.1) {\emph{Generation}};

    \node [realdata] (input) {Human data \\ (Query (q), Facet (f))};
    \node [highlight=gray!20, right=of input] (generate) {Generate clarifying questions (cq) };
    \node [generateddata, right=of generate] (pool) {Clarifying question pool};

    \node [process, highlight=teal!20, below=of pool] (scoring) {Filtering};
    \node [generateddata, below=1 cm of scoring] (output) {Ranked cqs};

    \node [highlight=red!20, left=1.5cm of output] (answers) {Generate answers (a)};
    \node [generateddata, left=1.5cm of answers, minimum height=2.7cm] (dataset) {Synthetic data \\ (cq,a)};

    \draw[->] (input) -- node[midway, above]{\textbf{q}} (generate);
    \draw[->] (generate) -- node[midway, above]{\textbf{cq}}  (pool);
    \draw[->] (pool) -- (scoring);
    \draw[->] (scoring) -- (output);
    \draw[->] (output) -- node[midway, above]{\textbf{(cq, f, q)}}  (answers);
    \draw[->] (answers) -- node[midway, above]{\textbf{(cq, a)}} (dataset);

    \node [phase, fit=(generate) (pool), label=above:Phase 1 - Clarifying question generation] {};
    \node [phase, fit=(scoring) (output), label={[verticaltext,yshift=-3.5mm, xshift=-2.15cm]right:Phase 2 - Question filtering}] {};  %
    \node [phase, fit=(answers), minimum height=1.43cm, label={[xshift=-0.1cm]above:Phase 3 - Answer generation}] {};

    \newcommand{\cube}[1]{%
    \coordinate (CenterPoint) at (0,0);
    \def\width{0.42cm};
    \def\height{0.42cm};
    \def\textborder{0.1cm};
    \def\xslant{0.1cm};
    \def\yslant{0.075cm};
    \def\rounding{0.2pt};
    \node[draw,
          minimum height  = \height,
          minimum width   = \width,
          text width      = {\width-2*\textborder},
          align           = center,
          fill            = #1!20,
          rounded corners = \rounding]
          at (CenterPoint) {}; 
    \draw [rounded corners = \rounding, fill=#1!10] %
        ($(CenterPoint) + (-\width/2. - 2*\rounding, \height/2.)$) -- %
        ($(CenterPoint) + (-\width/2. + \xslant - 2*\rounding, \height/2. + \yslant)$) -- %
        ($(CenterPoint) + (\width/2. + \xslant + 2*\rounding, \height/2. + \yslant)$) -- %
        ($(CenterPoint) + (\width/2. + 2*\rounding, \height/2.)$) -- %
        cycle;
    \draw [rounded corners = \rounding, fill=#1!30] %
        ($(CenterPoint) + (\width/2. + \xslant + 2*\rounding, \height/2. + \yslant)$) -- %
        ($(CenterPoint) + (\width/2. + 2*\rounding, \height/2.)$) -- %
        ($(CenterPoint) + (\width/2. + 2*\rounding, -\height/2.)$) -- %
        ($(CenterPoint) + (\width/2. + \xslant + 2*\rounding, -\height/2. + \yslant)$) -- %
        cycle;
        }        

\draw [black, rounded corners=2, thick] (-1.7,-10.4) rectangle (11,-7.6) ;

    \node at (9.4,-10.1) {\emph{Evaluation}};

	\node [below=3.325cm of dataset, xshift=-0.6cm, font=\small] {\textbf{CrowdLLM}};
	
	\node [coordinate, below=0.85cm of dataset] (aux1){};
	\draw [densely dotted] (dataset) -- (aux1);
	\node [coordinate, below=1.1cm of dataset] (aux2){};
	\draw [densely dotted] (aux2) -- ($(aux2)+(0,-1.1)$);
	
	\node [below=2.35cm of dataset] {$\textbf{(cq,a)}$};

	\begin{scope}[xshift=1.47cm,yshift=-8.31cm]
		\cube{blue}
    \end{scope}
	\begin{scope}[xshift=1.47cm,yshift=-9.01cm]
		\cube{blue}
    \end{scope}
	\begin{scope}[xshift=1.47cm,yshift=-9.71cm]
		\cube{blue}
    \end{scope}
        
    \node [coordinate] (qaPair) at (0.5cm, -9.01cm) {};
    \node [coordinate] (leftSplit) at (0.8cm, -9.01cm) {};
    \node [coordinate] (topJudge) at (1.4cm,-8.31) {};
    \node [coordinate] (midJudge) at (1.4cm,-9.01) {};
    \node [coordinate] (bottomJudge) at (1.4cm,-9.71) {};
        
    \draw (qaPair) -- (leftSplit);
    \draw [->,rounded corners=0.05cm] (leftSplit) -- ($(leftSplit)+(0.1,0)$) -- ($(topJudge)+(-0.4,0)$) -- ($(topJudge)+(-0.17,0)$) {};
    \draw [->,rounded corners=0.05cm] (leftSplit) -- ($(midJudge)+(-0.17,0)$) {};
    \draw [->,rounded corners=0.05cm] (leftSplit) -- ($(leftSplit)+(0.1,0)$) -- ($(bottomJudge)+(-0.4,0)$) -- ($(bottomJudge)+(-0.17,0)$) {};
        
	\node[draw,fill = teal!20,right=1.5cm of midJudge, minimum width=3.2cm, yshift=0.5cm, font=\small] (QEval) {Question evaluation};
	\node[draw,fill = red!20,right=1.5cm of midJudge, minimum width=3.2cm, yshift=-0.5cm, font=\small] (AEval) {Answer evaluation};
	
    \node [coordinate] (rightSplit) at (2.1cm, -9.01cm) {};
        
    \draw [->,rounded corners=0.05cm] ($(topJudge)+(0.3,0)$) -- ($(topJudge)+(0.5,0)$) -- (rightSplit) --($(rightSplit)+(0.3,0)$) -- ($(QEval.west)+(-0.2,0)$) -- (QEval.west) {};
    \draw [->,rounded corners=0.05cm] ($(topJudge)+(0.3,0)$) -- ($(topJudge)+(0.5,0)$) -- (rightSplit) --($(rightSplit)+(0.3,0)$) -- ($(AEval.west)+(-0.2,0)$) -- (AEval.west) {};
    \draw [rounded corners=0.05cm] ($(midJudge)+(0.3,0)$) -- ($(rightSplit)+(0.1,0)$) {};
    \draw [rounded corners=0.05cm] ($(bottomJudge)+(0.3,0)$) -- ($(bottomJudge)+(0.5,0)$) -- (rightSplit) --($(rightSplit)+(0.3,0)$) {};
        
    \newcommand{\report}[1]{%
        \coordinate (Starter) at (0,0);
        \path [fill=#1!40] (0,0) rectangle (0.09,0.3) ;
        \path [fill=#1!40] (0.1,0) rectangle (0.19,0.5) ;
        \path [fill=#1!40] (0.2,0) rectangle (0.29,0.2) ;        
        \path [fill=#1!40] (0.3,0) rectangle (0.39,0.35) ;  
        \path [fill=#1!40] (0.4,0) rectangle (0.49,0.15) ;          
    }
        
    \begin{scope}[xshift=6.75cm,yshift=-8.7cm]
		\report{blue}
    \end{scope}
    \begin{scope}[xshift=7.25cm,yshift=-8.7cm]
		\report{red}
    \end{scope}
    \begin{scope}[xshift=7.75cm,yshift=-8.7cm]
		\report{orange}
    \end{scope}
    \begin{scope}[xshift=8.25cm,yshift=-8.7cm]
		\report{magenta}
    \end{scope}
    \begin{scope}[xshift=8.75cm,yshift=-8.7cm]
		\report{teal}
    \end{scope}
    \begin{scope}[xshift=9.25cm,yshift=-8.7cm]
		\report{violet}
    \end{scope}

    \begin{scope}[xshift=6.75cm,yshift=-9.7cm]
		\report{brown}
    \end{scope}
    \begin{scope}[xshift=7.25cm,yshift=-9.7cm]
		\report{cyan}
    \end{scope}
    \begin{scope}[xshift=7.75cm,yshift=-9.7cm]
		\report{green}
    \end{scope}
    \begin{scope}[xshift=8.25cm,yshift=-9.7cm]
		\report{gray}
    \end{scope}
        
    \draw [->,rounded corners=0.05cm] (QEval.east) -- ($(QEval.east)+(0.62,0)$)  {};
    \draw [->,rounded corners=0.05cm] (AEval.east) -- ($(AEval.east)+(0.62,0)$)  {};                
\end{tikzpicture}
}
\vspace*{-3mm}
\caption{The AGENT-CQ framework for generating clarifying questions and simulating answers (top) and evaluating generated questions and answers (bottom).}
\label{fig:clarifying-questions-framework}
\end{figure}
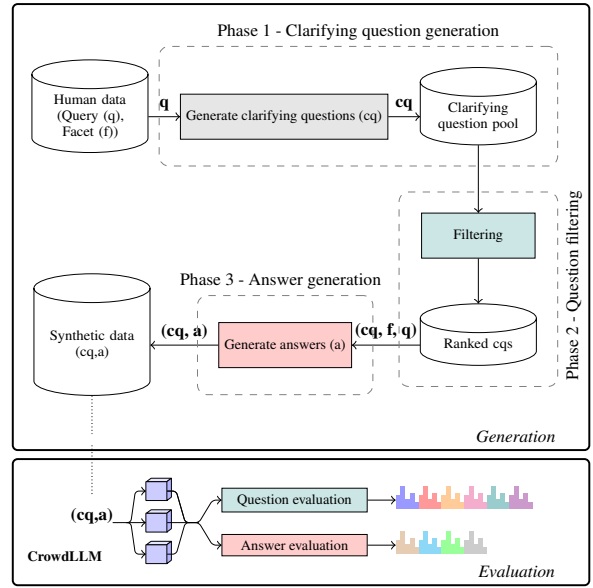

Existing methods for generating clarifying questions in \ac{CS} systems rely on manual curation by experts and template-based approaches~\citep{DBLP:journals/corr/abs-2009-11352,DBLP:conf/www/ZamaniDCBL20}: human experts craft clarifying questions, while using their ability to intuitively understand complex user intents and contextual nuances. 
While this method ensures high relevance and accuracy, it poses challenges for scalability in large-scale applications~\citep{DBLP:journals/air/DeriuROERAC21}. Moreover, human curators do not necessarily have deep knowledge about the topic of a conversation. In contrast, template-based methods employ pre-defined templates to automate the generation of clarifying questions, significantly enhancing scalability and efficiency. 
However, these methods often lack flexibility, leading to generic or less-diverse questions that could hurt the overall user interaction experience~\cite{yao-2012-semantics-based}. 

Recently, \acp{LLM} have shown capability in generating quality synthetic data for NLP and IR tasks, e.g., dialogue generation~\cite{ding2023enhancing}, document generation~\cite{askari2023expand}, and query generation~\cite{jeronymo2023inpars2}, with little work to explore the generation of clarifying questions. Therefore, we ask:
\emph{How can we use \acp{LLM} to generate synthetic data of clarifying questions in information-seeking dialogues, which can be used to train smaller models that are able to pose clarifying questions in dialogues more effectively?}

We propose \textbf{AGENT-CQ}, an end-to-end LLM-based framework for generating and evaluating clarifying questions. 
AGENT-CQ has two stages: \emph{generation} (top) and \emph{evaluation} (bottom); see Figure~\ref{fig:clarifying-questions-framework}. 
The generation stage has three main phases: question generation~(Phase~1), filtering~(Phase~2), and answer generation~(Phase~3). 
In Phase~1, we employ \acp{LLM} to generate and compare clarifying questions. 
In Phase~2, we filter out generated questions that do not meet certain quality criteria.
In Phase~3, we generate answers to the final set of questions by simulating system-user interactions. 

CrowdLLM is the second stage of AGENT-CQ; it is designed to assess the overall quality of the generated clarifying questions data.
It simulates a crowd of workers by employing three instances of an LLM to evaluate the generated questions and simulated answers, mimicking diversified human judgments. 
CrowdLLM is multidimensional assessing the question quality on seven quality metrics and four qualities for the answers.
Given the challenges of evaluating clarifying questions -- where human judgments can be time-consuming, costly, and subjective -- CrowdLLM offers a scalable and consistent alternative. We conduct an extensive human evaluation to ensure the quality and robustness of CrowLLM evaluation labels. \looseness=-1

Our experiments on the ClariQ dataset~\citep{aliannejadi-etal-2021-building} show that
\begin{enumerate*}[label=(\roman*)]
    \item CrowdLLM is highly effective at evaluating clarifying questions and answers, exhibiting strong inter-rater agreement across most dimensions with high correlation with human expert ratings; 
    \item GPT--Temp, our temperature-variation method, consistently outperforms other approaches, in terms of clarity, relevance, usefulness, and overall quality. Facet-based approaches (e.g., GPT-Facet) demonstrate high specificity but increased question complexity.
    
    \item LLM-generated clarifying questions, particularly those from GPT-Temp, enhance retrieval effectiveness across both BM25 and BERT models, consistently achieving higher NDCG scores than human-generated questions. 
\end{enumerate*} \looseness=-1

Our \textbf{main contributions} in this paper are:
\begin{enumerate*}[leftmargin=*,label=(C\arabic*),nosep]
    \item AGENT-CQ: A scalable methodology for generating and evaluating clarifying questions with LLMs.
    \item As part of AGENT-CQ, we also share a reliable evaluation framework (CrowdLLM) balancing scalability and accuracy in question assessment.
    \item A comparative analysis of LLM architectures for generating questions and simulating user responses.
\end{enumerate*} \looseness=-1

\section{AGENT-CQ Framework}
\label{section:method}

We introduce the two key stages of AGENT-CQ: a framework for generating clarifying questions and a framework for evaluation.

\subsection{AGENT-CQ: Generation framework}
AGENT-CQ's generation framework generates and scores clarifying questions using state-of-the-art LLMs in an end-to-end manner. The framework has three main phases; see Figure~\ref{fig:clarifying-questions-framework} (top).

\subsubsection{Question generation (Phase 1)}
Let $Q = \{q_1, q_2, \ldots, q_n\}$ be a set of $n$ initial user queries. For each $q_i \in Q$, we aim to generate a set of clarifying questions $C_i = \{c_{i1}, c_{i2}, \ldots, c_{im}\}$, where $m$ is the number of clarifying questions per set.
We define a question generation function:
\begin{equation}
G : q_i \rightarrow \{C_i^1, C_i^2, \ldots, C_i^k\}
\end{equation}
that generates $k$ sets of clarifying questions for query $q_i$.
We explore two prompt-based approaches (i.e., $p=2$).

\header{Facet-based approach} We adopt the approach of diverse query interpretation based on \citet{aliannejadi-etal-2021-building}, aiming to generate clarifying questions that address multiple interpretations of a given query. 
We introduce the facet-based method~\citep{DBLP:conf/www/ZamaniDCBL20,DBLP:conf/ictir/SekulicAC21}. Here, an LLM takes a query as input, then generates facets as a way of exploring the topic of the query and finally generates a clarifying from a query-facet pair. Algorithm~\ref{alg:facet-based} (Appendix~\ref{app:methodology-phase 1}) details the implementation of this approach. 

\header{Temperature-variation-based approach} This method generates diverse clarifying questions by systematically adjusting an LLM's temperature parameter. Starting from a low temperature and incrementing it over multiple iterations, it produces question sets with progressively increasing diversity. This approach implicitly explores various query facets, potentially uncovering different clarifications without explicit facet modeling. For a detailed description, see Algorithm~\ref{alg:temeperature-based} (Appendix~\ref{app:methodology-phase 1}).

\subsubsection{Question filtering (Phase 2)}
In this phase, we take two major characteristics of questions to reject low-quality questions. Preliminary experiments indicated that LLMs can sometimes generate questions that are not actually clarifying questions, or are not on the same topic as the user query. We define a function $S$ to filter out questions based on  relevance and clarification potential:
\begin{equation}
S(q_i, C_i^j) = \alpha \cdot R(q_i, C_i^j) + (1-\alpha) \cdot L(C_i^j),
\end{equation}
where
$R(q_i, C_i^j)$ is the relevance score;
$L(C_i^j)$ is the clarification score, evaluating the questions' potential to clarify user intent; and
$\alpha$ is a weighting parameter. We keep the top 10 ranked questions for each query in the collection for each LLM and experimental setup.

\subsubsection{User response simulation (Phase~3)}
In phase~3 of AGENT-CQ's generation stage, we simulate user responses to the ranked clarifying questions from Phase 2. 
Recent work has demonstrated the efficacy of simulated users as cost-effective proxies for real users in conversational systems~\citep{yoon-etal-2024-evaluating,zhao-2024-let,DBLP:conf/wsdm/Abbasiantaeb0KA24}. 
Using this insight, we employ an approach that takes LLM as a simulator for generating answers to system-generated clarifying questions.
We introduce a \textit{parameterized-user simulation} approach, inspired by \citet{DBLP:journals/tist/SekulicAC24}. This method incorporates user characteristics~($U$) in the simulation, to generate diverse and realistic answers (details see Algorithm~\ref{alg:parameterized_response} Appendix~\ref{app:methodology-response}). Our parameterized function is defined as: 
\begin{equation}
a_{ij} = A_p(q_i, u_i, c_{ij}, U),
\end{equation}
where $q_i$ is the original query, $u_i$ is the user information need, $c_{ij}$ is the clarifying question, and $U$ is the set of user characteristics.
$A_p$ extends the basic non-parameterized method\footnote{\scriptsize{Defined as:
$a_{ij} = A_Np(q_i, u_i, c_{ij})$}} by incorporating user characteristics $U$, primarily verbosity, which controls the response length, detail and revealment probability used to determine the likelihood of disclosing the true user information need. 

\subsection{AGENT-CQ: Evaluation framework}
Next, we detail CrowdLLM, AGENT-CQ's evaluation framework.
CrowdLLM is a multi-LLM and multi-dimensional framework evaluating the generated questions and simulated responses using the scalability of LLMs to simulate a crowd of evaluators, with validation from human experts.

\header{Multi-LLM evaluation}
CrowdLLM employs an LLM-as-a-judge approach \citep{DBLP:conf/nips/ZhengC00WZL0LXZ23}, in an ensemble of LLM instances with varying temperature settings. 
We hypothesize that with varying temperatures different LLM instances will bring different angles to the evaluation. 
This design simulates the setup used with crowdsourced workers, crucial for comprehensive assessment in NLP tasks. 
Each LLM instance evaluates questions and answers on multiple aspects using a 10-point scale for questions and pairwise comparison between LLM and human answers. 
To validate CrowdLLM's performance, we incorporate human expert assessment. 

\header{Evaluation metrics}
Evaluation in CrowdLLM is based on distinct sets of metrics for clarifying questions and simulated answers, drawn from prior work on conversational information seeking and general conversational systems \citep{DBLP:conf/www/ZamaniDCBL20,aliannejadi-etal-2021-building,DBLP:conf/acl/DaumeR18,siro-etal-2024-context,Siro-UsatCRS}.
For clarifying questions, we assess clarification potential, on-topic relevance, specificity, usefulness, clarity, and question complexity. 
Simulated answers are evaluated on relevance, usefulness, naturalness, and overall quality. \footnote{\scriptsize{Definition of the aspects in Appendix~\ref{app:methodology-metrics}}}

Details about our experimental setup, implementation details and prompts used are included in Appendix~\ref{app:methodology-impl} and \ref{app:prompts}.

\section{Reliability of CrowdLLM}
In this section, we study the reliability of our evaluation framework~(CrowdLLM) from multiple angles.

\begin{table*}[t!]
\centering
\begin{tabular}{l cc cc cc cc cc}
    \toprule
        \multirow{2}{*}{\textbf{Aspects}} & \multicolumn{2}{c}{\textbf{GPT-Baseline}} & \multicolumn{2}{c}{\textbf{GPT-Facet}} & \multicolumn{2}{c}{\textbf{GPT-Temp}} & \multicolumn{2}{c}{\textbf{Llama 3.1}} & \multicolumn{2}{c}{\textbf{H-Gen}} \\
    \cmidrule(r){2-3} \cmidrule(r){4-5} \cmidrule(r){6-7} \cmidrule(r){8-9} \cmidrule(r){10-11}
     & ICC & W-$\kappa$ & ICC & W-$\kappa$ & ICC & W-$\kappa$ & ICC & W-$\kappa$ & ICC & W-$\kappa$ \\
    \midrule
    Clarification & 0.96 &0.87  & 0.95 &0.85  &0.85  & 0.72 & 0.97 & 0.89 & 0.97  & 0.89 \\
    Clarity & 0.90 & 0.79 & 0.80 & 0.67 & 0.81 & 0.77 & 0.94 & 0.84 & 0.95 & 0.87 \\
    On-topic & 0.93 &0.81  & 0.87 & 0.78 & 0.87 & 0.82 & 0.93 & 0.86 & 0.96 & 0.88 \\
    Question-C & 0.86 & 0.76 & 0.93 & 0.84 & 0.87 &0.80  & 0.94 & 0.85 & 0.78 & 0.73 \\
    Specificity & 0.92 & 0.80 & 0.83 & 0.66 & 0.80 & 0.66 & 0.92 & 0.79 & 0.96 &0.87  \\
    Usefulness &0.94  & 0.84 & 0.93 & 0.82 & 0.92 & 0.78 & 0.97 & 0.89 & 0.97 & 0.90 \\
    Overall-quality & 0.92 &0.81  &0.88  &0.82  & 0.75 & 0.68 & 0.94 & 0.85 & 0.95 &0.88  \\
    \bottomrule
\end{tabular}
\caption{CrowdLLM ICC and weighted $\kappa$ (W-$\kappa$) agreement scores for different prompting strategies across models including human generated clarifying questions (H-Gen). Question-C denotes question complexity.}
\label{tab:IAA}
\end{table*}

\subsection{Clarifying questions}
In Table~\ref{tab:IAA}, we report the inter-annotator agreement (IAA) among the LLM instances (i.e., GPT-4o) using \ac{ICC} and weighted $\kappa$~\cite{cohen-1968-weighted}. We further assess the agreement of CrowdLLM with human evaluators to inform on the quality of the evaluations. Because of the large number of generated questions by each model and the associated costs, our human evaluators assess a sample of 200 questions from each model and rank them based on preference.
For each user information need and initial request, human evaluators see the clarifying questions generated by each system and rank them from most helpful (Rank 1) to least helpful (Rank 5) and give a justification for the least helpful clarifying question. For CrowdLLM evaluations we rank the questions based on overall quality score using Tukey's HSD post-hoc test for pairwise comparisons.

\textbf{Internal agreement.} 
\emph{CrowdLLM demonstrates moderately consistent performance across most aspects of question quality.} Overall quality shows notable variability ($ICC= 0.75$, $\kappa= 0.68$), lower than other aspects. This suggests that temperature variation in LLM instances leads to diverse perspectives on holistic question evaluation. Specificity ($ICC=0.80$, $\kappa= 0.66$) and clarity ($ICC= 0.81$, $\kappa= 0.77$) show lower inter-rater agreement. Higher-temperature instances interpret abstract details as specific and rate subtly ambiguous questions as clear, while lower-temperature instances focus on explicit information and apply stricter clarity standards. Despite these variations, high agreement across most aspects indicates CrowdLLM provides a robust method for evaluating clarifying question quality.

\begin{figure}[t]
    \centering
    \includegraphics[width=1\linewidth]{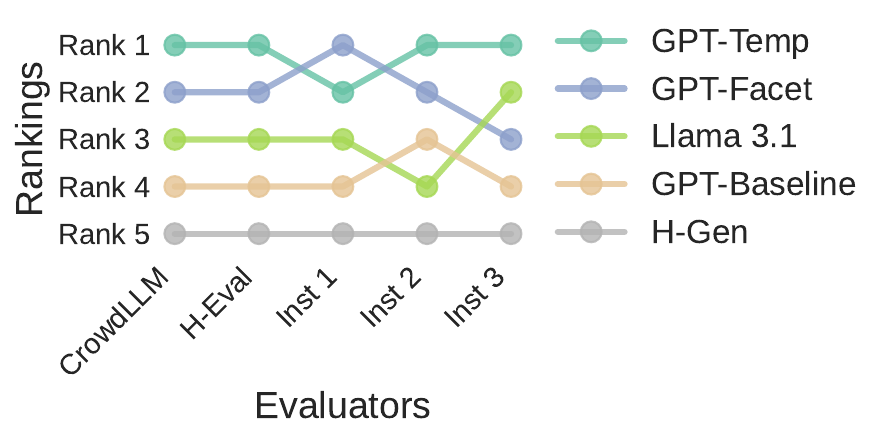}
    \caption{Rankings of the question sets from different systems by different evaluators.}
    \label{fig:qn-rank}
\end{figure}

\textbf{External agreement.}
\emph{CrowdLLM evaluations strongly align with human evaluation results~(H-Eval), confirming its effectiveness~(Figure~\ref{fig:qn-rank}).}  Both approaches consistently rank the GPT-Temp question set as most helpful and human-generated questions (H-Gen) as least helpful, with H-Gen receiving an average rank of 4 out of 5 in human evaluation (lower is better). CrowdLLM instances show ranking variations: Instance 1 prioritizes GPT-Facet $>$ GPT-Temp $>$ Llama 3.1; Instance 2 mirrors aggregate rankings with GPT-Baseline $>$ Llama 3.1; Instance 3 ranks Llama 3.1 second, above GPT-Facet and GPT-Baseline.
These inconsistencies highlight the value of using multiple instances to produce robust, human-aligned assessments. Despite the mid-rank differences, all evaluation approaches consistently rank GPT-Temp or GPT-Facet first and H-Gen last.

\subsection{Simulated answers}
We assessed CrowdLLM's reliability in evaluating generated answers, comparing it with human evaluators (H-Eval). Evaluators judged which answer in presented pairs was better for naturalness, relevance, usefulness, and overall quality, with an option to rate them equal. Table \ref{tab:inter-rater-reliability} presents inter-rater reliability results for this comparative evaluation.

\textbf{Internal agreement.}
\emph{CrowdLLM shows high internal consistency.} Naturalness has highest agreement ($\kappa = 0.81$, 89\% agreement), indicating near-perfect consensus. Relevance ($\kappa = 0.68$, 86\% Ag.) and overall quality ($\kappa = 0.71$, 79\% Ag.) show substantial agreement. Usefulness, while substantial, has lowest scores ($\kappa = 0.62$, 73\% Ag.).

\begin{table}[ht]
\centering
\setlength{\tabcolsep}{1.5mm}
\begin{tabular}{lccc}
\toprule

& \textbf{Fleiss' $\kappa$} & \textbf{\% Ag.}  & \textbf{\% H-Ag.} \\
\midrule
Naturalness & 0.81 & 89\% & 53\% \\
Relevance & 0.68 & 86\% & 73\% \\
Usefulness & 0.62 & 73\% & 68\% \\
Overall-quality & 0.71 & 79\% & 75\% \\
\bottomrule
\end{tabular}
\caption{Inter-rater reliability measures for CrowdLLM and Human evaluators in terms of Fleiss' $\kappa$, annotator agreement percentage, and Human-CrowdLLM agreement (H-Ag.).}
\label{tab:inter-rater-reliability}
\end{table}

\begin{figure}[t]
    \centering
    \includegraphics[clip, trim=0mm 10mm 0mm 12mm,width=0.75\linewidth]{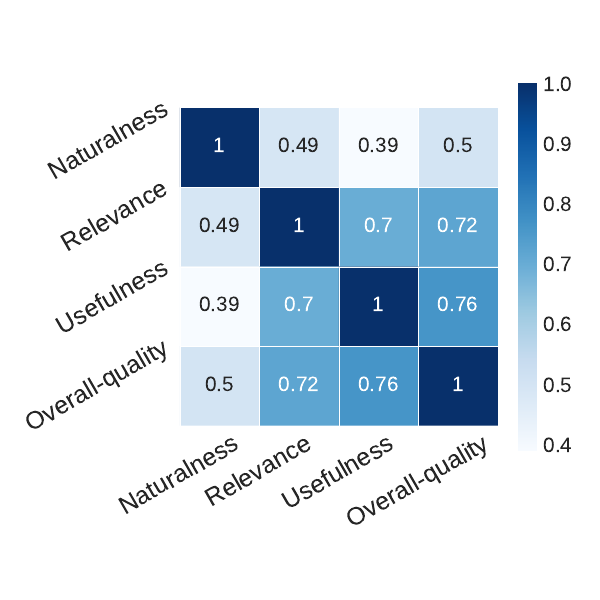}
    \caption{Spearman's $\rho$ correlations between answer evaluation aspects.}
    \label{fig:ans-corr}
\end{figure}

\textbf{External agreement.} \emph{CrowdLLM and human evaluators align strongly in overall quality (75\%) and relevance (73\%).} Usefulness agreement was moderate (68\%). \citet{DBLP:conf/sigir/SiroAR24} showed humans better evaluate usefulness of system responses than LLMs, supporting our findings. Naturalness had lowest agreement (53\%), despite high internal consistency in both methods. 
This may be due to LLMs' bias towards their generated answers, while humans better discern between human and LLM-generated responses~\citep{DBLP:conf/naacl/SahaLCBWL24}.

\subsection{Effectiveness of evaluation aspects}
\emph{Which aspects most influence perceived quality?}
Table~\ref{tab:qn-corr} shows correlations between question aspects and overall quality. 
Usefulness has the strongest association ($\tau = 0.80$, $\rho = 0.90$), followed by clarification ($\tau = 0.76$, $\rho = 0.87$), clarity ($\tau = 0.75$, $\rho = 0.85$), and on-topic relevance ($\tau = 0.71$, $\rho = 0.81$). 
Specificity shows moderate to strong correlation ($\tau = 0.63$, $\rho = 0.73$); query-complexity has negligible impact ($\tau = 0.07$, $\rho = 0.08$). 
Thus, usefulness, clarification, clarity, and topical relevance are key determinants of perceived question quality in CrowdLLM evaluation.

Figure~\ref{fig:ans-corr} shows Spearman's $\rho$ correlations between answer aspects and overall quality. Usefulness ($\rho = 0.76$) and relevance ($\rho = 0.72$) correlate strongest with overall quality, indicating their critical role in perceived answer quality. Naturalness shows moderate correlation ($\rho = 0.50$), suggesting less impact. Strong correlations between relevance and usefulness ($\rho = 0.70$) highlights their interconnectedness in high-quality answers. Naturalness correlates less with relevance ($\rho = 0.49$) and usefulness ($\rho = 0.39$); it captures a distinct dimension of answer quality.

In general,  \emph{CrowdLLM evaluations reveal that usefulness and relevance are critical for both questions and answers. Questions benefit significantly from clarity and clarification, while answers balance usefulness, relevance, and naturalness.}

\begin{table}[t]
\centering
\begin{tabular}{lcc}
\toprule
\textbf{Aspect}      & \textbf{Kendall's $\tau$} & \textbf{Spearman's $\rho$} \\ 
\midrule
Clarification             & 0.76                     & 0.87                       \\ 
Clarity                 & 0.75                     & 0.85                       \\ 
On-Topic               & 0.71                     & 0.81                       \\ 
Question-C       & 0.07                    & 0.08                     \\ 
Specificity               & 0.63                    & 0.73                       \\ 
Usefulness                 & 0.80                   & 0.90                       \\ 
\bottomrule
\end{tabular}
\caption{Kendall's $\tau$ and Spearman's $\rho$ correlations of CrowdLLM question evaluation aspects with overall quality. Question-C denotes question complexity}
\label{tab:qn-corr}
\end{table}

\section{Evaluation of Generated Clarifying Questions}

We perform a comprehensive analysis to assess the quality of data generated by AGENT-CQ.

\subsection{Clarifying question evaluation}
We conduct an analysis to explore the characteristics of the generated clarifying questions. 
We focus on three areas: identifying recurring question patterns~\citep{DBLP:conf/chiir/BraslavskiSAD17}, categorizing questions based on their intent~\citep{DBLP:conf/chiir/BraslavskiSAD17,DBLP:conf/www/ZamaniDCBL20}, and classifying the expected response types. We employ a hierarchical matching system to analyze linguistic patterns and key phrases. 
Categorization uses LLMs to capture subtle distinctions between types such as disambiguation and information seeking as shown in Table~\ref{tab:clarifying_categories} (Appendix~\ref{appendix:analysis-techniques}). We classify response types into Yes/No, Multiple Choice, Open-ended, and Factual using a rule-based approach. Detailed analysis techniques in~Appendix~\ref{appendix:analysis-techniques}.

\textbf{Question length and readability analysis.} Table~\ref{tab:qn-length-readability} shows that human questions are concise (9.71 words) and simple (5th-grade level). LLM outputs vary: GPT-Facet generates complex, lengthy questions (college-level, 23.53 words), Llama 3.1 generates variable-length high school-level questions, and GPT-baseline closely matches human question length but with higher complexity. 
There is a consistent gap in LLMs' ability to replicate the brevity and simplicity of human-written questions.

\textbf{Question categories.}
Table~\ref{tab:qn-categories} shows that all models except Llama 3.1 favor \textit{preference identification} questions, with GPT-Facet leading at 74.00\%. Llama 3.1 has a more balanced distribution between \textit{preference identification} (47.20\%) and \textit{information gathering} (41.00\%), and the highest disambiguation rate (10.20\%). Human-generated questions have the highest confirmation rate (17.91\%). Comparison questions are consistently low ($<$1.61\%) across all approaches.

\textbf{Question patterns and response types.}
Table~\ref{tab:qn-patterns} and~\ref{tab:qn-presponses} (Appendix~\ref{appendix:analysis-findings}) reveal distinct question patterns and response types across human and LLM outputs. Humans show greater pattern diversity, favoring ``Would you like'' (21.17\%) and ``Do you need/want/have'' (17.74\%), with a strong preference for Yes/No responses (80.68\%). In contrast, ``Are you X'' dominates GPT-Facet (75.80\%) and GPT-Temp (50.80\%), aligning with their preference for Multiple Choice responses ($\approx$73\%). Llama 3.1 most closely mirrors human diversity. GPT-Baseline shows unique tendencies, preferring ``Do you need/want/have'' (20.32\%) and Yes/No responses (70.22\%). Some patterns (e.g., ``What specific'') are almost exclusively LLM-generated, highlighting significant differences in question formulation between humans and LLMs.

Generally, \emph{LLMs exhibit model-specific tendencies in generating clarifying questions, often diverging from human patterns w.r.t.\ question structure, expected response type, category focus, length, and complexity, highlighting the challenges in replicating natural human question-asking behavior.}

\begin{table}[t]
\centering
\setlength{\tabcolsep}{1mm}
\resizebox{\columnwidth}{!}{%
\begin{tabular}{lccccc}
\toprule
\textbf{Model} & \textbf{Pref.} & \textbf{Info.} & \textbf{Disamb.} & \textbf{Conf.} & \textbf{Comp.} \\
\midrule
Llama 3.1 & 47.20 & 41.00 & 10.20 & \phantom{0}0.60 & 1.00 \\
GPT-Baseline & 73.64 & 15.29 & \phantom{0}2.41 & \phantom{0}7.04 & 1.61 \\
GPT-Facet & 74.00 & 18.00 & \phantom{0}6.20 & \phantom{0}0.60 & 1.20 \\
GPT-Temp & 66.80 & 14.40 & 16.40 & \phantom{0}1.60 & 0.80 \\
Human & 64.39 & 10.66 & \phantom{0}6.84 & 17.91 & 0.20 \\
\bottomrule
\end{tabular}
}
\caption{Percentage of question categories for different smodels. Columns: Pref: Preference identification, Info: Information seeking, Disamb: Disambiguation, Conf: Confirmation, Comp: Comparison.}
\label{tab:qn-categories}
\end{table}

\subsection{Quality analysis of clarifying questions using CrowdLLM}

\begin{figure}[!t]
    \centering
    \includegraphics[width=1\columnwidth]{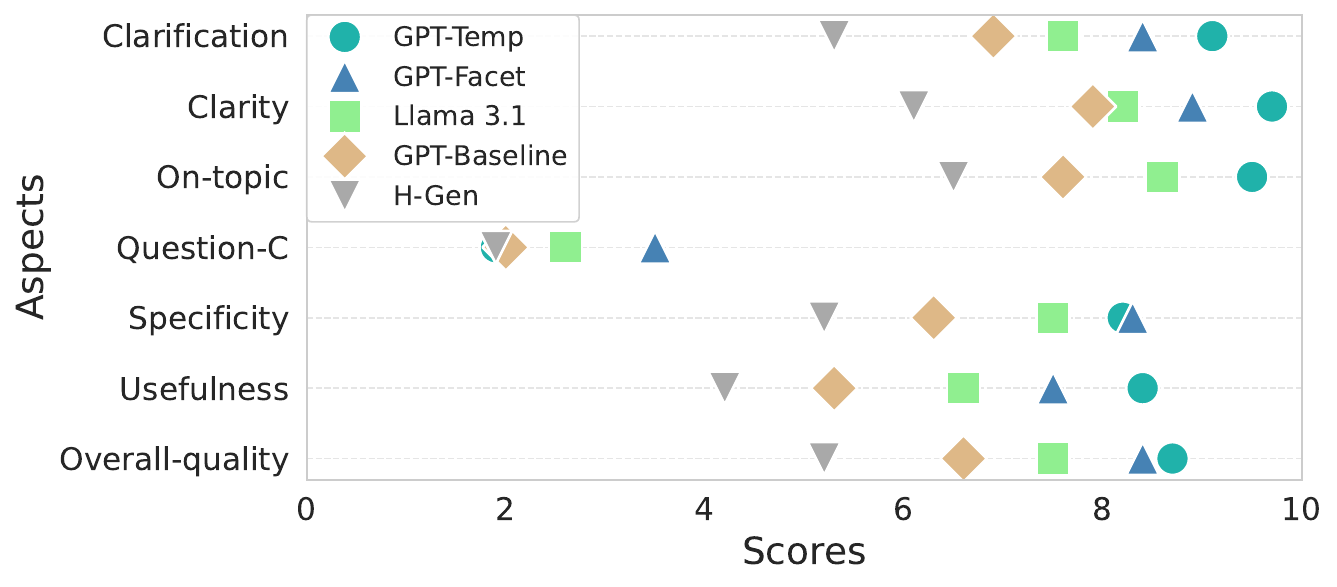}
    \caption{Mean question quality scores evaluated by CrowdLLM across all aspects for different models. H-Gen: human-generated questions.}
    \label{fig:mean-rating}
\end{figure}

We assess clarifying questions from different models and prompting strategies across seven aspects: \textit{clarification}, \textit{on-topic}, \textit{specificity}, \textit{usefulness}, \textit{clarity}, \textit{query complexity}, and \textit{overall quality}. Figure~\ref{fig:mean-rating} shows mean scores across all aspects per model. We use one-way ANOVA with post-hoc Tukey's HSD for statistical analysis ($p < 0.05$). Detailed explanations and test suitability are in Appendix~\ref{app:statistical_methods}. \looseness=-1

\begin{table}[!t]
\centering
\vspace*{2mm}
\setlength{\tabcolsep}{1.5mm}
\begin{tabular}{l@{~}cc rr@{~}l}
\toprule
\textbf{Model} & \textbf{Mean} & \textbf{Std} & \multicolumn{3}{c}{\textbf{Flesch-Rea. (Kin.)}} \\
\midrule
Llama3.1 & 19.02 & 9.26 &\phantom{X}& 53.77 &(9th) \\
GPT-Baseline & 10.72 & 2.15 && 61.10 & (7th) \\
GPT-Facet & 23.53 & 4.98 && 35.58 & (14th) \\
GPT-Temp & 15.68 & 3.33 && 52.74 & (9th) \\
Human & \phantom{0}9.71 & 2.48 && 75.99 & (5th) \\
\bottomrule
\end{tabular}
\caption{Question length statistics and readability scores. Length differences are statistically significant, $p<0.05$; Kin:Flesch-Kincaid grade level in brackets.}
\label{tab:qn-length-readability}
\end{table}

GPT-Temp consistently outperforms other approaches across most aspects (Figure~\ref{fig:mean-rating}), surpassing GPT-Baseline in usefulness (mean difference $= 3.781$, $p < 0.001$). Facet-based models (GPT-Facet and Llama 3.1) show improvements over the baseline, with GPT-Facet often ranking second. GPT-Facet excels in specificity, significantly outperforming GPT-Baseline, as it generates specific facets before producing targeted clarifying questions. 
Thus, facet-based approaches enhance specificity but generate more complex questions (GPT-Facet: 3.5) than GPT-Temp and human-generated questions (both 1.9), aligning with our Fleisch readability and Kincaid analysis.

Human-generated questions score lowest across most aspects, except for complexity. GPT-Temp significantly outperforms human questions in usefulness (8.4 vs.\ 4.2, $p < 0.001$), challenging assumptions about human expertise in question formulation. LLMs' superior performance can be attributed to their knowledge and consistent optimization for specific criteria. 
Recent research shows LLMs favor their own content~\citep{liu2024llms}; our human evaluators corroborate CrowdLLM's results by also ranking human questions as least helpful. This alignment between human and LLM evaluations validates our conclusions and suggests that CrowdLLM is not biased in this case.

\begin{table*}[!t]
\centering
\begin{tabular}{l rrr rrr}
\toprule
 \multirow{2}{*}{\textbf{Aspects}} & \multicolumn{3}{c}{\textbf{CrowdLLM} } & \multicolumn{3}{c}{\textbf{H-Eval} }  \\
\cmidrule(r){2-4} \cmidrule(r){5-7} 
                 & Human  & LLM-answer  & Tie & Human  & LLM-answer  & Tie   \\
\midrule
Relevance        & 37.34           & 37.15              & 19.71            & 32.7                  & 36.6                  & 30.7                \\
Usefulness       & 38.26         & 41.86             & 14.47          & 36.6                & 38.6           & 24.8                \\
Naturalness      & 37.74                 &  55.16$^{*}$               & 6.82                  & 32.0                & 34.0                   & 34.0                \\
Overall-quality  & 45.52                  & 53.17$^{*}$                 & 1.82                  & 39.9                & 41.8$^{*}$                   & 18.3                \\
\bottomrule
\end{tabular}
\caption{Percentage of pairwise comparisons won by each model and ties, as evaluated by CrowdLLM and human evaluators (H-Eval). * indicates statistical significance (trinomial test, $p < 0.05$).}
\label{tab:win_tie_comparison}
\end{table*}

\begin{table}[!t]
\centering
\setlength{\tabcolsep}{1mm}
\resizebox{\columnwidth}{!}{%
\begin{tabular}{ll@{~}cccc}
\toprule
& \textbf{Questions} & \textbf{NDCG@1} & \textbf{NDCG@5} & \textbf{NDCG@10} \\
\midrule
\multirow{4}{*}{\rotatebox[origin=c]{90}{\textbf{BM25}}} 
& GPT-Baseline (L) & 0.180 & 0.187 & 0.209 \\
& GPT-Temp (L) & \textbf{0.225} & 0.199 & 0.214\\
& Human (H) & 0.201  & \textbf{0.221} &  \textbf{0.246}\\
& Human (L) &0.173 & 0.193 & 0.215\\
\midrule 
\multirow{4}{*}{\rotatebox[origin=c]{90}{\textbf{BERT}}} 
& GPT-Baseline (L) & 0.283 & 0.294 & \textbf{0.303} \\
& GPT-Temp (L) &\textbf{0.312}  & \textbf{0.296} & 0.301\\
& Human (H) & 0.307 & 0.288& 0.301\\
& Human (L) & 0.267& 0.259 &0.277\\
\bottomrule
\end{tabular}
}
\caption{BM25 and BERT retrieval and ranking performance with different clarifying questions. In brackets answer source: (H) - human answers and (L) - LLM answers.}
\label{tab:summary_distribution}
\label{tab:bm25-ret}
\end{table}

In summary, \emph{LLM-generated clarifying questions, particularly from GPT-Temp, outperform human-generated ones across most quality aspects. Temperature variation is highly effective. GPT-Temp's strong performance and low complexity make it ideal for general-purpose clarification tasks. Facet-based approaches enhance specificity but increase complexity; they best fit specialized domains requiring detailed clarifications.}

\subsection{Evaluation of simulated answers}
We conducted pairwise comparisons of 200 answer pairs across four aspects: relevance, usefulness, naturalness, and overall quality. Each pair was evaluated by three human workers and three CrowdLLM instances. We define a win for a model when at least two out of three (human or LLM) evaluators agree that the model's answer is superior; if the majority rates the answers as equal, we have a tie.

LLM responses are longer (mean 13.21 vs.\ 8.19 words) and more variable (std dev.\ 8.06 vs.\ 4.36) than human-generated ones. Table~\ref{tab:win_tie_comparison} shows that LLM answers perform comparably to human answers in relevance and usefulness, demonstrating our approach's success in generating contextually appropriate and valuable responses. Naturalness assessments yield intriguing results: human evaluators slightly favor LLM answers (34\% vs.\ 32\%), while CrowdLLM shows a stronger preference (55.16\% vs.\ 37.74\%), suggesting potential bias in automated evaluation systems. Overall quality marginally favors LLM answers with statistical significance, contrasting with previous non-parametric simulations where human evaluators consistently preferred human answers~\citep{DBLP:journals/tist/SekulicAC24}. 

Overall, \emph{our parametric approach generates LLM-simulated answers that closely match or slightly outperform human answers across key aspects, contrasting with previous non-parametric simulations and demonstrating successful capture of real user response diversity.}

\section{Retrieval Performance Comparison}

Following \citet{aliannejadi21building}, we evaluate the impact of  clarifying questions on document retrieval performance. Our methodology simulates a typical conversational search scenario: a user initiates a search, the system poses a clarifying question, and the user provides an answer. The retrieval system then uses this additional information to retrieve an updated set of documents. We hypothesize that higher-quality clarifying questions and their corresponding answers lead to improved post-QA retrieval performance.

Table~\ref{tab:bm25-ret} presents the retrieval performance using BM25 and BERT models with clarifying questions and their corresponding answers. GPT-Temp-generated clarifying questions significantly enhance retrieval effectiveness for both the BM25 and BERT models.  For BM25, GPT-Temp questions achieve the highest NDCG@1 (0.225), demonstrating superior performance in top-rank retrieval. In BERT-based retrieval, GPT-Temp leads with NDCG@1 of 0.312 and NDCG@5 of 0.296. This performance aligns with our earlier findings, where both evaluators ranked GPT-Temp questions as the most helpful; its ability to generate precise, contextually relevant questions directly translates to improved retrieval outcomes.

\noindent%
Human-generated questions with human answers perform better in BM25 retrieval at NDCG@5 (0.221) and NDCG@10 (0.246). This effectiveness likely stems from two factors: humans' tendency to use terms overlapping with the original query, enhancing lexical matching, and the overall quality of human answers, contributing to improved ranking quality in term-based retrieval. However, when human questions are paired with LLM-generated answers, performance declines across both retrieval models. 
This contrasts with our finding that LLM-simulated answers are often indistinguishable from human answers in quality assessments. Suggesting that while our parametric approach successfully mimics human-like responses, it may not fully capture the nuanced interactions between clarifying questions and answers crucial for retrieval tasks. \looseness=-1

Generally, \emph{GPT-Temp, excel in generating high-quality clarifying questions, enhancing top-ranked retrieval across models. However, BM25 benefits more from precise human-generated content, while BERT leverages contextually rich LLM questions effectively. This highlights the need for LLMs to better integrate query-relevant terms and context for optimal cross-model retrieval performance.}

\section{Related Work}

\textbf{LLM-based \ac{CS} systems.}
Conversational search is an interactive paradigm where users engage in a dialogue with a search system~\cite{DBLP:conf/chiir/RadlinskiC17}. 
In conversational search systems, LLMs enhance the search experience through query understanding, retrieving relevant documents, and generating clear responses~\cite{DBLP:journals/corr/abs-2404-18424,DBLP:conf/emnlp/DengLC0LC23,DBLP:conf/sigir/JokoCRV0H24}.
LLMs have been adopted to simulate users and their interactions with the system, reducing the need for human resources ~\cite{DBLP:journals/corr/abs-2305-13614,DBLP:journals/corr/abs-2407-13166,Ren2024BASESLW,DBLP:journals/corr/abs-2402-05746,DBLP:conf/acl/NiuWCS024}. E.g., \citet{DBLP:conf/wsdm/Abbasiantaeb0KA24}
simulate teacher-student interactions in a conversational setting. 
\citet{DBLP:journals/tist/SekulicAC24} focus on evaluating query clarification via an LLM-based user simulator. 

Our work focuses on query clarification, i.e., the process of refining or elaborating on a user's initial search query or question to better understand their intent~\cite{AliannejadiZCC19,DBLP:conf/www/ZamaniDCBL20}. 
Prior work on the role of clarifying questions in conversational search recognizes their potential to enhance search quality~\cite{DBLP:conf/cikm/AliannejadiAZK021,DBLP:conf/sigir/TavakoliTZSS22,DBLP:conf/www/0002SARL24} and the user experience~\cite{DBLP:conf/ecir/SekulicAC21,DBLP:conf/sigir/ZamaniMCLDBCD20}.
How LLMs can benefit the task remains underexplored. 
Our work evaluates LLMs on query clarification at the clarifying question and user response levels, assessing their ability to generate effective questions and responses. \looseness=-1

\textbf{Evaluation of generated content.}
Traditional automated metrics for evaluating generated content like BLEU \citep{DBLP:conf/acl/PapineniRWZ02} and ROUGE \citep{lin-2004-rouge} often correlate poorly with human judgments for open-ended text generation \citep{DBLP:conf/emnlp/LiuLSNCP16}. 
Human evaluation is the gold standard but time-consuming and costly to scale~\citep{DBLP:conf/acl/KiritchenkoM17}. 
Newer metrics (USR, \citeauthor{DBLP:conf/acl/MehriE20}, \citeyear{DBLP:conf/acl/MehriE20}; BLEURT, \citeauthor{DBLP:conf/acl/SellamDP20}, \citeyear{DBLP:conf/acl/SellamDP20}; 
BERTScore, \citeauthor{DBLP:conf/iclr/ZhangKWWA20}, \citeyear{DBLP:conf/iclr/ZhangKWWA20}) use pre-trained language models to improve correlation with human judgments. 
Multi-dimensional evaluation frameworks have also emerged~\citep{DBLP:conf/acl/DouFKSC22,DBLP:journals/corr/abs-2104-05361}.
Recent research explores using LLMs as evaluators for natural language tasks~\citep{DBLP:conf/eamt/KocmiF23,DBLP:conf/emnlp/LiuIXWXZ23}.
\citet{DBLP:conf/nips/ZhengC00WZL0LXZ23} explore ``LLM-as-a-judge'' for chat assistants.
\citet{lin-chen-2023-llm} introduce LLM-Eval, a multi-dimensional evaluation method for open-domain conversations.

CrowdLLM shares similarities with recent LLM-based evaluation techniques \citep{DBLP:conf/emnlp/LiuIXWXZ23,DBLP:conf/nips/ZhengC00WZL0LXZ23} and multi-dimensional frameworks \citep{DBLP:conf/acl/DouFKSC22}. However, it uniquely employs multiple LLM instances to simulate diverse evaluators, addressing scalability issues of human evaluation. 
CrowdLLM also incorporates a second tier where humans assess the reliability of LLM-generated evaluations, combining automated efficiency with human judgment accuracy. \looseness=-1

\section{Conclusion}
We introduce AGENT-CQ, a framework for generating and evaluating clarifying questions and answers in \ac{CS} systems. Our study reveals that GPT-Temp consistently outperforms other methods in generating high-quality clarifying questions. Surprisingly, human-generated questions, despite lower quality ratings, excelled in term-based retrieval at NDCG@5 and 10. LLM-simulated answers, while matching human answers in quality assessments, underperformed in retrieval tasks when paired with human questions. CrowdLLM, our evaluation framework, showed general alignment with expert assessments but demonstrated potential biases towards LLM-generated content. 
Future work should explore on enhancing LLMs' ability to generate retrieval-effective questions and answers, improving the integration of LLM-generated content with existing retrieval models. 
\looseness=-1

\clearpage
\section*{Limitations}
Our study, using the ClariQ dataset, demonstrates the potential of LLMs in generating effective clarifying questions, but has several limitations. While we employed various LLM models and prompting strategies to broaden our question range, the dataset may not fully represent real-world query diversity, potentially limiting result generalizability. The  nature of LLMs challenges full understanding of their decision-making process, which we addressed through detailed analyses of question patterns, response types, and quality aspects.
Additionally, while we’ve made significant efforts in prompt design, the effectiveness of our method still inherently relies on the quality of the prompts for both generation and filtering.
Moreover, LLMs may not consistently capture all relevant query aspects, especially for complex or niche topics.

\section*{Ethical Considerations}

We carefully considered ethical implications throughout our study. To address potential biases in LLMs, we compared their outputs with human-generated questions from the ClariQ dataset and conducted thorough evaluations using both CrowdLLM and human assessors. This approach helped identify and mitigate potential unfairness in question generation across different topics. We prioritized privacy by using only publicly available datasets and ensuring that no personally identifiable information was processed or generated. In our human evaluation process, we ensured fair compensation , adhering to ethical guidelines for crowdsourced work. 

While our study shows LLMs' superior performance in many aspects of question generation, we recognize that there are critical domains where they may not be applicable, emphasizing the continued importance of human involvement. This is crucial in fields requiring specialized knowledge, ethical decision-making, or handling sensitive information, where human expertise remains irreplaceable.

To address these limitations and ethical concerns, future work should focus on several key areas. Domain-specific adaptation is crucial, exploring how our approach can be tailored for specialized fields requiring expert knowledge. This could involve developing domain-adapted LLMs or targeted prompts for more precise clarifying questions in specific contexts.
Further research is needed to identify and mitigate biases in LLM-generated questions, ensuring fairness across diverse topics and user groups. This work is essential for creating more equitable and universally applicable systems.
Extending our approach to multi-turn interactions presents another challenge. Future studies should explore how LLMs can maintain coherence and relevance across multiple rounds of clarification, simulating more natural, in-depth conversations. This advancement would enhance LLM capability in complex information-seeking scenarios, bringing AI-assisted search closer to human-like interaction patterns.

\section*{Acknowledgements}
This research was supported by the Dreams Lab, a collaboration between Huawei Finland, the University of Amsterdam, and the Vrije Universiteit Amsterdam, by the Hybrid Intelligence Center, a 10-year program funded by the Dutch Ministry of Education, Culture and Science through the Netherlands Organisation for Scientific Research, https://hybrid-intelligence-centre.nl, by project LESSEN with project number NWA.1389.20.183 of the research program NWA ORC 2020/21, which is (partly) financed by the Dutch Research Council (NWO), and by the FINDHR (Fairness and Intersectional Non-Discrimination in Human Recommendation) project that received funding from the European Union’s Horizon Europe research and innovation program under grant agreement No 101070212.

All content represents the opinion of the authors, which is not necessarily shared or endorsed by their respective employers and/or sponsors.

\bibliography{anthology,references}
\bibliographystyle{acl_natbib}

\clearpage
\appendix
\section{Additional Methodology Details}\label{app:methodology}
In this section we give additional details on the implementation of AGENT-CQ.

\subsection{Clarifying Question Generation Algorithms} \label{app:methodology-phase 1}
The two alternative methods used for clarifying question generation in Phase~1 of AGENT-CQ are detailed in Algorithm~\ref{alg:facet-based} and~\ref{alg:temeperature-based}.

\begin{algorithm}[h!]
\caption{Facet-based clarifying question generation}
\label{alg:facet-based}
\begin{algorithmic}[1]
\Require Query $q_i$
\Ensure Set of clarifying questions $C_i$
\State $F_i \gets \phi(q_i)$ \Comment{Generate facets}
\State $C_i \gets \{\}$
\For{each facet $f_{ij} \in F_i$}
    \State $c_{ij} \gets \psi(q_i, f_{ij})$ \Comment{Generate questions}
    \State $C_i \gets C_i \cup \{c_{ij}\}$
\EndFor \\
\Return $C_i$
\end{algorithmic}
\end{algorithm}

\begin{algorithm}[h!]
\caption{Temperature-variation-based clarifying question generation. 
}
\label{alg:temeperature-based}
\begin{algorithmic}[1]
\Require Query $q_i$, Temperature variations $k$
\Ensure Set of clarifying questions $C_i$
    \State $C_i \gets \{\}$
    \For{$j \gets 1$ \textbf{to} $k$}
        \State $\tau \gets \min(0.9, 0.5 + (j-1) * 0.1)$ 
        \\ \Comment{Update LLM's temperature $\tau$}
        \State $c_{ij} \gets \psi(q_i, \tau)$ \Comment{Generate questions}
        \State $C_i \gets C_i \cup \{c_i^j\}$
    \EndFor
    \State \Return $C_i$
\end{algorithmic}
\end{algorithm}

\subsection{User Response Simulation Algorithm}\label{app:methodology-response}
Algorithm~\ref{alg:parameterized_response} describes the parameterized answer simulation approach by LLM.

\begin{algorithm}[h!]
\caption{Parameterized User Response Generation}
\label{alg:parameterized_response}
\begin{algorithmic}[1]
\Require Query $q_i$, Facet $f_i$, Set of clarifying question $C_i$, User characteristics $U$
\Ensure Set of parameterized responses $A_i$
\State $A_i \gets {}$
\For{each $c_{ij} \in C_i$}
\State $\gets \text{ConstructParameterizedPrompt}(q_i, f_i, c_{ij}, U)$
\State $a_{ij} \gets \psi(q_i, f_i, c_{ij}, U)$ \Comment{Generate response}
\State $A_i \gets A_i \cup {a_{ij}}$
\EndFor
\State \Return $A_i$
\end{algorithmic}
\end{algorithm}

The $\texttt{ConstructParameterizedPrompt}$ function generates a structured prompt by incorporating the original query $q_i$, user information need $u_i$, and clarifying question $c_{ij}$, along with verbosity level and reveal probability randomly selected from the user characteristics set $U$. This approach generates a wider range of responses, better reflecting real-world user behavior diversity. It also provides a richer dataset for training and evaluating conversational search systems, enabling systematic study of user characteristics' impact on system performance.

\subsection{CrowdLLM Question and Answer Evaluation Metrics} \label{app:methodology-metrics}
\paragraph{Question quality metrics.}
CrowdLLM evaluated the quality of the generated clarifying questions from a multidimensional perspective capturing the following quality aspects:

\begin{enumerate}[leftmargin=*,nosep]
    \item Clarification: Assesses how well the question seeks to understand the original query without introducing unrelated topics.
    \item On-topic: Measures the question's direct relation to the subject matter of the original query.
    \item Specificity: Evaluates the question's focus on particular aspects of the query rather than being general.
    \item Usefulness: Gauges how much answering the question would improve the response to the original query.
    \item Clarity: This measure evaluates how easily understood and unambiguous the clarifying question is from the user's perspective.
    \item Question complexity: This aspect examines whether the clarifying question introduces technical terms, specialized concepts, or requires domain-specific knowledge not present in the original query.
    \item Overall quality: Assesses the overall quality of the question based on the above metrics
\end{enumerate}

\paragraph{Answer quality metrics.}
Similar to questions answers we also evaluated from a multidimensional perspective on the following three metrics and overall quality.

\begin{enumerate}
\item Relevance: How directly the user's answer addresses the system's clarifying question.
\item Usefulness: The value of the user's answer in clarifying their original information need.
\item Naturalness: The human-like quality and conversational tone of the user's response.
\item Overall quality: Holistic assessment of the answer's effectiveness in aiding system understanding.
\end{enumerate}

\subsection{Implementation Details} \label{app:methodology-impl}

\subsubsection{Dataset} We use an existing question clarification dataset, named ClariQ~\citep{aliannejadi-etal-2021-building}. ClariQ is one of the most widely-used question clarification dataset and aligns well with our setting. Each data sample in ClariQ includes a topic originated from the TREC Web Track
2009–2012~\citep{DBLP:conf/trec/ClarkeCS09} that represents an initial user query. These topics can be further divided into multiple facets that capture the user's true intent. For each facet, a set of manually collected clarifying questions are provided which helps the system better understand the underlying user intention.
Subsequently, user responses are collected for each clarifying question, providing insights to the corresponding facet. Notably, the ground-truth retrieved documents are also attached given each topic-facet pair. 
Specifically, in our setting, we reuse the queries from ClariQ but prompt \acp{LLM} to generate diverse clarifying questions and simulate user responses. This diversification allows us to better simulate real-world scenarios where users may have different perspectives or require more specific information. Overall, it consists 198 topics with 891 different facets and over 8k questions, with 9.49 terms on average per question.

\subsubsection{Models}
For our experiments we use GPT~\citep{Brown-gpt} and Llama models~\citep{llama-Touvron} in our framework. For question generation, we primarily employ GPT-based models. In the facet-based method, we use a hybrid approach: GPT-3.5 generates query facets, which are then fed to Llama for question generation, as Llama alone was ineffective in facet generation. The same generation model is used in the filtering stage to evaluate and select the most appropriate questions. We used the 8B variant of Llama-3.1. For simulating user responses, we rely on GPT-3.5 due to its versatility in generating diverse and contextually appropriate answers. Our CrowdLLM evaluation framework uses GPT-4o~\citep{gpt4-report} as the base model, leveraging its advanced capabilities for assessing question and answer quality.

\subsubsection{ Hyperparameters}
Our framework employs various hyperparameters, carefully chosen to balance performance and diversity:
\begin{enumerate}
\item Question generation:
\begin{itemize}
\item Temperature variation: We use temperatures ranging from 0.5 to 0.9, incrementing by 0.1. We set $n$\_$ sets = 3$.
\item Facet-based approach: Temperature is set to 0.7, top\_p = 0.95, for Llama: top\_k = 50 and max\_length = 1024.
 \item Baseline: A fixed temperature of 0.7 is used to generate 10 questions for each query.
\end{itemize}
\item Question filtering:
\begin{itemize}
    \item  We set $\alpha = 0.4$ in the filtering stage to balance relevance and clarification potential of the selected questions.
    \item Temperature is set to 0.7.
\end{itemize}
\item User simulation: 
\begin{itemize}
    \item Verbosity: 10--60 tokens
    \item Cooperativeness: reveal probabilities 0.0--0.9
    \item Answer generation: temperature = 0.7, top\_p = 0.98, frequency\_penalty = 0.5, presence\_penalty = 0.2
\end{itemize}
These parameters simulate diverse user behaviors while maintaining coherent responses.
\item CrowdLLM evaluation: We use three GPT-4 instances to simulate diverse human judgments:
\begin{itemize}
    \item Conservative judge (temperature 0.2): Produces predictable, focused judgments, simulating a strict evaluator.
    \item Balanced judge (temperature 0.5): Provides a mix of creativity and focus, representing a typical evaluator.
    \item Creative judge (temperature 0.7): Generates more exploratory judgments, simulating a lenient evaluator.
\end{itemize}

\end{enumerate}

The selection of these hyperparameters was based on:
\begin{itemize*}
\item Extensive experimentation with various setups to optimize performance.
\item Analysis of output quality and diversity across different parameter combinations.
\item Alignment with observed patterns in human evaluation behaviors from prior crowdsourcing studies.
\end{itemize*}

\section{Prompts}\label{app:prompts}
In this section, we list the prompts used in different prompting strategies and stages of AGENT-CQ.

\subsubsection{Facet-based Prompt}
\begin{lstlisting}
For the user query: '{query}'
Generate a list of 40 diverse facets that this query might be addressing.
This query represents multiple user information needs. Generate diverse facets to capture these varied needs.
Ensure each facet is unique and explores different aspects or interpretations of the query. Avoid repetition and strive for a wide range of perspectives in your facets.


\end{lstlisting}

\begin{lstlisting}
For the user query: '{query}'
And considering this specific facet: '{facet}'
Generate a clarifying question that addresses this facet and helps to better understand the user's specific information need.
Use diverse language and question structure to formulate the questions.
\end{lstlisting}

\subsection{Temperature-variation prompt}

\begin{lstlisting}
for i in range(n_sets):
    For the user query: '{query}'

    Generate a set of 10 clarifying questions. The goal is to better understand the user's specific information need.
    
    This query represents multiple user information needs. Generate diverse clarifying questions to capture these varied needs. 
    Ensure each question is unique and explores different aspects or interpretations of the query. Avoid repetition and strive for a wide range of perspectives in your questions.
    
    IMPORTANT GUIDELINES:
    1. Each question should aim to clarify a different aspect of the user's intent or information need.
    2. Ensure all questions are unique. Do not repeat questions.
    3. Focus on questions that will help narrow down or specify the user's request.
    4. Consider potential ambiguities or multiple interpretations of the query.
\end{lstlisting}

\subsection{Scoring and filtering prompt}

\begin{lstlisting}
Evaluate the following question for the user query: '{query}'
Question: "{question}"
Consider these aspects:
    1. Clarification: How well does this question help to better understand the user's original query?
    2. On Topic: To what degree does this question directly relate to the subject matter of the user's original query?

Provide a score (0-10) for each aspect and a brief explanation.
\end{lstlisting}

\subsubsection{User response simulation prompt}

\begin{lstlisting}
You are a user who initially made this request: '{query}'. 
Your actual information need is: '{facet}'. 
Respond to the clarifying question based on this information need. 

Your verbosity level is {verbosity_level}. 
Your reveal probability is {reveal_probability:.2f}.    
Keep your response short, ideally under {verbosity["max_tokens"]} tokens.

Remember: Your answer should not include any additional information that is not part of your actual information need ('{facet}'). 

        
\end{lstlisting}

\subsection{CrowdLLM prompt}
Below is an example of CrowdLLM prompt for question complexity. Other metrics follow the same prompt except for the definition of the metric. Each metric is evaluated independently to avoid bias from previous metric rating. Overall quality followed a slightly different approach, apart from having access to the query and system clarifying question, it also included the ratings from the other six metrics in order to ground the overall quality on these metrics.
\begin{lstlisting}
As a user, you are evaluating the complexity of the system's clarifying question in relation to your original query.

Definition:
    - Question Complexity: The degree to which the clarifying question introduces technical terms, specialized concepts, or requires domain-specific knowledge not present in the original query.

Scale:
    1-10, where 1 is very simple (uses only general terms and concepts) and 10 is highly complex (introduces specialized terminology or concepts).

Your original query: "{original_query}"
System's clarifying question: "{system_question}"

Evaluate the complexity of the system's question compared to your original query. Consider:
    1. Does it introduce technical terms or jargon not present in the original query?
    2. Does it require specialized knowledge that might not be evident from the original query?
\end{lstlisting}

\begin{lstlisting}
As a user, you are providing an overall evaluation of the system's clarifying question, taking into account your ratings from other aspects.

Definition:
    - Overall Quality: Your comprehensive assessment of how well the system's clarifying question helps you get a better response to your original query, considering clarity, relevance, specificity, and usefulness.

Scale:
    1-10, where 1 is the lowest quality and 10 is the highest quality.

Your original query: "{original_query}"
System's clarifying question: "{system_question}"

Your rating from the other metrics: {other_ratings}

Consider these ratings and provide an overall evaluation of the system's clarifying question quality. Explain your reasoning, referencing your other metric ratings.

\end{lstlisting}

\section{Human Evaluation}
To quantify the effectiveness of our evaluation framework (CrowdLLM) we conducted human evaluation to assess both the question and answer quality. We employed the so called Master crowdworkers from Amazon Mechanical Turk from the US with an approval rate of more 95\% in over 10000 HITS. Each HIT was done by 3 workers and they were paid \$8.5 per hour.

Different from CrowdLLM which evaluated each question on six dimensions then gave overall quality, humans assessed the questions based on preference. This is because of the large number of the questions and associated costs. In each HIT a worker was shown the initial user request and five generated clarifying questions for the query from each system: Llama 3.1, GPT-Facet, GPT-Temp, GPT-Baseline and Human question. They were tasked to rank the questions from the most helpful (Rank 1) to the least helpful (Rank 5) using a drag and drop option. To avoid position bias, where a system's question is always placed at the top, we uniformly randomized the order of the questions at each HIT so that each system question was placed at the top in 20\% of the HITS. A total of 1000 questions were assessed, 200 from each system.

Similar to CrowdLLM, humans assessed a pair of answers on three dimensions: Relevance, usefulness, naturalness and on overall answer quality. The comparison was between human answers and LLM-simulated answers to human clarifying questions. We used human clarifying questions because they already had human generated answers and the dataset is well known and has been used in several research, therefore allowing us to do a comparison with our LLM-generated answers, thus avoiding collection of new human answers.
At each HIT the workers were presented with a user information need which we call facet in our case, the initial user query, the human clarifying question and two answers; one from human and one from an LLM. Their task was to assess this pair of answers on the four dimensions and choose which of the answer was more relevant, useful, natural and overall of high quality. If both answers were of the same quality then an option of 'Equal' could be selected. Similarly the order of the answers were randomly swapped. The workers assessed 100 answer pairs and in total 200 answers were assessed.

\section{Supplementary Results and Analyses}\label{app:results}

\subsection{Analyses}\label{appendix:analysis-techniques}

\paragraph{Question categories} We developed a classification framework for clarifying questions based on \citet{DBLP:conf/www/ZamaniDCBL20} and \citet{DBLP:conf/chiir/BraslavskiSAD17}. Table~\ref{tab:clarifying_categories} presents each question type with descriptions and examples, including user questions (UQ) and corresponding clarifying questions (CQ) to illustrate their application in real conversations. This taxonomy provides a robust framework for comparing clarification strategies across various \acp{LLM} in conversational information seeking.

Initial rule-based classification attempts proved inadequate for capturing nuances. For instance, ``Did you mean the book or the movie?'' could be categorized as disambiguation or information gathering, depending on context. To address this, we employed GPT-3.5 for categorization, leveraging its context-awareness to select the most appropriate category. This approach enabled more accurate classification, especially for questions requiring nuanced interpretation or potentially fitting multiple categories.

\paragraph{Question patterns} We developed a systematic approach to identify and classify question patterns using a hierarchical matching system. This process analyzes linguistic structure and key phrases, starting with primary question words (e.g., ``What'', ``How'', ``Are you'') and then examining subsequent words for more specific patterns. For example, ``What specific'' and ``What kind of'' are categorized differently from general ``What'' questions. ``How'' questions are differentiated based on inquiries about methods, duration, or extent. We also consider compound structures like ``Are you looking for'' or ``Do you need'', which are common in clarifying questions. Implementation uses a combination of regular expressions and string matching algorithms, balancing flexibility in pattern recognition with consistency in categorization. This approach enables nuanced analysis of how different prompting strategies formulate questions.

\paragraph{Elicited response types} We classified expected response types of clarifying questions into four categories: Yes/No, Multiple Choice, Open-ended, and Factual. Yes/No questions are identified by auxiliary verb initiation (e.g., ``Are'', ``Is'', ``Do''). Multiple Choice questions contain explicit options or suggest selection from a limited set. Factual questions use specific question words (e.g., ``When'', ``Where'', ``Who'') seeking concise information. Open-ended questions, not fitting other categories, typically invite elaboration. We implemented this classification using regular expressions and conditional logic.

To ensure accuracy, particularly for edge cases, we followed the automated classification with a manual review process. This combined approach allowed us to systematically analyze large volumes of clarifying questions while maintaining high classification accuracy. By examining patterns, categories, and response types, we gained insights into how different models and prompting strategies influence the structure and intent of clarifying questions generated by language models in conversational information seeking contexts.

\subsection{Findings} \label{appendix:analysis-findings}

Table~\ref{tab:clarifying_categories} shows the questions categories used to categorize the generated clarifying questions. In Table~\ref{tab:qn-patterns} and \ref{tab:qn-presponses} we show distributions of the question patterns and responses in detail.

\begin{table*}[t!]
    \centering
    \begin{tabular}{p{0.22\textwidth} p{0.32\textwidth} p{0.38\textwidth}}
        \toprule
        \textbf{Category} & \textbf{Description} & \textbf{Example} \\
        \midrule
        Disambiguation~\citep{DBLP:conf/www/ZamaniDCBL20}
        & Addresses queries that are ambiguous and could refer to different concepts or entities. 
        & UQ: I'm looking for information on Java \newline CQ: Are you referring to Java the programming language, Java the island, or Java coffee? \\
        \midrule
        Preference Identification~\citep{DBLP:conf/www/ZamaniDCBL20} 
        & Clarifies the user's specific preferences, including personal, spatial, temporal, or purpose-related information. 
        & UQ: I want to buy a new laptop \newline CQ: What will be the primary use of this laptop? Gaming, work, or general use? \\
        \midrule
        Information Gathering~\citep{DBLP:conf/www/ZamaniDCBL20,DBLP:conf/chiir/BraslavskiSAD17} 
        & Seeks additional details, verifications, or narrows down broad topics. 
        & UQ: Tell me about artificial intelligence \newline CQ: Which aspect of artificial intelligence are you most interested in learning about: machine learning, neural networks, or natural language processing? \\
        \midrule        Comparison~\citep{DBLP:conf/www/ZamaniDCBL20,DBLP:conf/chiir/BraslavskiSAD17} 
        & Involves comparing entities or options to aid decision-making. 
        & UQ: I'm researching electric cars \newline CQ: Would you like to compare the range, performance, or price of different electric car models?\\
        \midrule        Confirmation~\citep{DBLP:conf/www/ZamaniDCBL20,DBLP:conf/chiir/BraslavskiSAD17} 
        & Questions that seek to verify or confirm previously provided information or assumptions. 
        & UQ: I need a new phone \newline CQ: Are you specifically looking for a smartphone, or would you consider other types of mobile phones? \\
        \midrule
        General~\citep{DBLP:conf/chiir/BraslavskiSAD17} 
        & Broad questions that prompt for additional details or elaboration on a topic.
        & UQ: I want to start a business \newline CQ: Can you provide more details about your business idea and what stage of planning you're in? \\
        \bottomrule
    \end{tabular}
    \caption{Clarifying question categories and examples. UQ stands for User Query, which represents an example of a typical user question. CQ stands for Clarifying Question, which shows how a system might respond to the UQ by asking for more specific or relevant information.}
    \label{tab:clarifying_categories}    
\end{table*}

\begin{table*}[t!]
\centering
\setlength{\tabcolsep}{0.5mm}
\begin{tabular}{l@{}cccccc}
\toprule
\textbf{Pattern} & \textbf{Llama3.1 (\%)} & \textbf{GPT-Baseline (\%)} & \textbf{GPT-Facet (\%)} & \textbf{GPT-Temp (\%)} & \textbf{H-Gen (\%)} \\
\midrule
Other & 29.00 & 15.90 & \phantom{0}1.40 & 11.60 & 29.44 \\
are you X & 22.60 & 37.63 & 75.80 & 50.80 & 29.64 \\
what specific & 20.60 & \phantom{0}2.62 & 20.40 & \phantom{0}5.00 & \phantom{0}0.00 \\
do you need/want/have & \phantom{0}6.80 & 20.32 & \phantom{0}0.00 & 22.80 & 17.74 \\
would you like & \phantom{0}4.60 & 18.11 & \phantom{0}0.00 & \phantom{0}4.80 & 21.17 \\
how X & \phantom{0}2.00 & \phantom{0}0.60 & \phantom{0}1.60 & \phantom{0}0.00 & \phantom{0}0.20 \\
are you looking for X & \phantom{0}1.40 & \phantom{0}0.00 & \phantom{0}0.00 & \phantom{0}0.20 & \phantom{0}0.00 \\
which specific & \phantom{0}1.20 & \phantom{0}0.00 & \phantom{0}0.80 & \phantom{0}0.00 & \phantom{0}0.00 \\
is there & \phantom{0}0.40 & \phantom{0}2.82 & \phantom{0}0.00 & \phantom{0}2.80 & \phantom{0}0.81 \\

\bottomrule
\end{tabular}
\caption{Results showing the percentage distribution od question patterns generated by various models including human questions (H-Gen).}
\label{tab:qn-patterns}
\end{table*}

\begin{table*}[t!]
\centering
\setlength{\tabcolsep}{1mm}
\begin{tabular}{lccccc}
\toprule
\textbf{Response type} & \textbf{Llama3.1 (\%)} & \textbf{GPT-Baseline (\%)} & \textbf{GPT-Facet (\%)} & \textbf{GPT-Temp (\%)} & \textbf{H-Gen (\%)} \\
\midrule
Multiple Choice & 41.40 & 22.33 & 73.20 & 73.00 & 10.46 \\
Open-ended & 37.20 & \phantom{0}6.84 & \phantom{0}8.80 & \phantom{0}3.40 & \phantom{0}4.02 \\
Yes/No & 16.00 & 70.22 & 17.80 & 22.60 & 80.68 \\
Factual & \phantom{0}5.40 & \phantom{0}0.60 & \phantom{0}0.20 & \phantom{0}1.00 & \phantom{0}4.83 \\
\bottomrule
\end{tabular}
\caption{Results showing the percentage distribution of response types elicited by the generated questions from various models including human questions (H-Gen). }
\label{tab:qn-presponses}
\end{table*}

\subsection{Statistical Methods for Question Quality Analysis}
\label{app:statistical_methods}
In our analysis of question quality across different models and prompting strategies, we employed two distinct statistical approaches to ensure robust and comprehensive results:

\header{Per-Question Analysis}
For the per-question analysis, we used one-way ANOVA (Analysis of Variance) with post-hoc Tukey's HSD (Honestly Significant Difference) test \cite{DBLP:journals/tit/Tukey66}.
\begin{itemize}[leftmargin=*,nosep]
\item \textbf{One-way ANOVA:} This test was chosen to compare the means of question quality scores across multiple groups (models/strategies). It is suitable for comparing three or more groups and assumes normality of distributions and homogeneity of variances.
\item \textbf{Tukey's HSD:} As a post-hoc test, Tukey's HSD was used to determine which specific groups differed from each other. It controls for Type I error and is appropriate for pairwise comparisons when sample sizes are large and approximately equal.
\end{itemize}

\noindent%
This approach is well-suited for our per-question analysis due to the potentially large sample sizes and the need to compare multiple groups simultaneously.

\section{Data Statistics and Sample Generated Data by Various Models}
In Table~\ref{data-statistics} we report the number of questions generated by different systems and the overall filtered questions. Table~\ref{tab:multiple-queries-models} shows sample clarifying questions generated by various models for each user query. Table~\ref{tab:human-llm-comparison} shows a sample of user responses to the human-generated clarifying questions by humans and an LLM. 

Given that GPT-Temp demonstrated superior performance in question quality, we utilize it to generate clarifying questions for all queries in the ClariQ collection. To maintain consistency with the original dataset, we match the number of clarifying questions per facet as present in the original data. This approach ensures a fair retrieval comparison while leveraging the enhanced question generation capabilities of GPT-Temp.
For comparative analysis, we also generate a dataset using a GPT-Baseline model. This allows us to assess the relative performance gains achieved by GPT-Temp and provides insights into the effectiveness of different prompting strategies in generating clarifying questions for \ac{CS} systems.

\begin{table}[!t]
\centering
\begin{tabular}{lccc}
\toprule
\textbf{Model} & \textbf{ Queries}  & \textbf{Generated} & \textbf{Filtered} \\ \midrule
Llama3.1 & 50 & 2000 & 500\\ 
GPT-Baseline & 50 & 500 & 500\\ 
GPT-Facet & 50 & 2000& 500 \\ 
GPT-Temp & 50 & 1500 & 500 \\ 
H-Gen& 50 & 1200 & 500\\ 
Total & 50 & 7200& 2500 \\ 
\bottomrule
\end{tabular}
\caption{Number of queries used in the experiments and total questions generated and filtered for each model.}
\label{data-statistics}
\end{table}

\begin{table*}[!t]
\centering
\resizebox{\textwidth}{!}{%
\begin{tabular}{p{0.14\textwidth} p{0.17\textwidth} p{0.17\textwidth} p{0.17\textwidth} p{0.17\textwidth} p{0.17\textwidth}}
\toprule
\textbf{Query} & \textbf{Llama3.1} & \textbf{GPT-Baseline} & \textbf{GPT-Facet} & \textbf{GPT-Temp} & \textbf{H-Gen} \\
\midrule
Tell me about atypical squamous cells 
& What specific information about atypical squamous cells are you looking for?
& Is there a specific type of testing or screening you're curious about? 
& What specific cytopathological techniques are you interested in regarding the diagnosis of atypical squamous cells? 
& Are you interested in the treatment options available if atypical squamous cells are detected?
& Would you like to know if there is a cure?  \\
\hline
Tell me information about computer programming.
& What specific aspect of the history of computer programming are you interested in?
& Do you want to know about programming tools and software?
& What specific problem-solving strategies or techniques in programming are you interested in learning more about?
& Are you looking for beginner resources, advanced topics, or something in between?
& Are you interested in a coding bootcamp	\\
\hline
Tell me about defender 
& Are you interested in learning more about the features and capabilities of defender as a cybersecurity tool?
& Are you referring to a specific type of defender, like in sports? 
& Are you looking for an explanation of defensive coding practices in general, or are you interested in specific techniques or examples of how they are implemented in programming? 
& Are you referring to a specific type of defender, such as a legal defender, a sports defender, or a cybersecurity defender?
& Are you interested in the television series defender\\
\hline
How to cure angular cheilitis
& What specific over-the-counter treatments are you looking for to cure angular cheilitis?
& What symptoms are you experiencing with angular cheilitis?
& Are you looking for home remedies or professional medical treatments to manage angular cheilitis in children?
& Would you prefer natural remedies, over-the-counter options, or prescription medications?
& Are you looking for the definition of angular cheilitis \\
\hline

\end{tabular}
}
\caption{Clarifying questions generated by different models for various queries}
\label{tab:multiple-queries-models}
\end{table*}

\begin{table*}[h]
\centering
\resizebox{\textwidth}{!}{%
\begin{tabular}{p{0.15\textwidth} p{0.2\textwidth} p{0.2\textwidth} p{0.2\textwidth} p{0.2\textwidth}}
\toprule
\textbf{Query} & \textbf{User-information-need} & \textbf{Clarifying-question} & \textbf{Human-answer} & \textbf{LLM-answer} \\
\midrule
Tell me about atypical squamous cells  & What do atypical squamous cells mean on a pap smear test? & Would you like to know if there is a cure?  & no i want to know what it means when atypical squamous cells appear on a pap smear test & Not specifically. Just curious about their meaning. \\
\hline
Tell me information about computer programming. & What type of careers are there for computer programmers? & Are you interested in a coding bootcamp & no i want to know what career options programmers have & I'm curious about the various job roles within computer programming. \\
\hline
Tell me about defender & Find information on the Land Rover Defender sport-utility vehicle. & Are you interested in the television series defender 3 & no i am looking for information about the land rover defender & Not exactly, I'm interested in a specific vehicle model. \\
\hline
How to cure angular cheilitis & What home remedies are there for angular cheilitis? & are you looking for the definition of angular cheilitis & no home remedies & I'm more interested in finding home remedies for angular cheilitis. \\
\hline

\end{tabular}
}
\caption{Comparison of Human and LLM answers to Human-Generated clarifying questions}
\label{tab:human-llm-comparison}
\end{table*}

\end{document}